# A New Pruning Method for Solving Decision Trees and Game Trees

**Prakash P. Shenoy**
School of Business
University of Kansas
Summerfield Hall
Lawrence, KS 66045-2003, USA
p-shenoy@ukans.edu

## Abstract

The main goal of this paper is to describe a new pruning method for solving decision trees and game trees. The pruning method for decision trees suggests a slight variant of decision trees that we call scenario trees. In scenario trees, we do not need a conditional probability for each edge emanating from a chance node. Instead, we require a joint probability for each path from the root node to a leaf node. We compare the pruning method to the traditional rollback method for decision trees and game trees. For problems that require Bayesian revision of probabilities, a scenario tree representation with the pruning method is more efficient than a decision tree representation with the rollback method. For game trees, the pruning method is more efficient than the rollback method.

## 1 INTRODUCTION

The main goal of this paper is to describe a new pruning method for solving decision trees and game trees. Decision trees and game trees are graphical methods for representing Bayesian decision problems.

The decision tree representation method was formulated by Raiffa and Schlaifer [1961] based on von Neumann and Morgenstern's [1944] extensive form game representation of $n$-person games. In extensive form games, information available to the decision makers is encoded using information sets. In decision trees, however, information about uncertainties is encoded by sequencing of the chance and decision variables. One consequence of this encoding is that Bayesian revision of probability models may be required before decision problems can be represented as decision trees. Decision trees are solved using the rollback method in which decision and chance nodes in the tree are pruned recursively starting from nodes adjacent to the leaves. Chance nodes are pruned by averaging the utilities at the ends of its edges, and decision nodes are pruned by maximizing the utilities at the ends of its edges.

Recently, Shenoy [1993b] has suggested using information sets to encode information constraints in decision problems. The resulting representation is called a game tree. Game trees generalize decision trees in the sense that a decision tree is a game tree in which all information sets are singletons. The rollback method of decision trees generalizes to game trees [Shenoy 1993b]. An advantage of game trees over decision trees is that if we have a Bayesian network model [Pearl 1988] for the uncertainties in a problem, then no preprocessing is ever required before a problem can be represented as a game tree. This is because information sets allow us to sequence the chance variables in the game tree as in the Bayesian network model.

In this paper, we propose a new pruning method to solve decision trees and game trees. The pruning method for decision trees is as follows. First we compute the path probabilities for each path from the root node to a leaf node by simply multiplying all the probabilities on the path. Next, we compute the weighted utility for each leaf node by multiplying the utility and the path probability. Next we prune all nodes starting from the nodes adjacent to leaf nodes and proceeding toward the root node as in the rollback method. Chance nodes are pruned by adding the weighted utilities at the ends of its edges, and decision nodes are pruned by maximizing the weighted utilities at the ends of its edges. This pruning method generalizes to game trees.

Notice that in the pruning method for decision trees, the probabilities on the chance edges are used only to compute the path probabilities (they are not used to prune chance nodes). The path probabilities are in fact joint probabilities. In decision problems that require a Bayesian revision of probabilities, we can avoid computation of the conditionals by simply computing the joint probability distribution and using the joint probabilities as path probabilities. Since conditionals are not necessary, we do not have to include them in the representation. We call this slight variant of decision trees scenario trees.

For decision problems that involves Bayesian revision of probabilities, a scenario tree representation solved using the pruning method requires less computation than a decision tree representation solved using the rollback method. For solving game trees, the pruning method requires less computation than the rollback method.

An outline of the remainder of the paper is as follows.

**Table I**. The Physician's Utility Function For All Act-State Pairs

| Physician's Utilities (υ) | | Events | | | |
|---|---|---|---|---|---|
| | | Has pathological state ($p$) | | No pathological state ($\sim p$) | |
| | | Has disease ($d$) | No disease ($\sim d$) | Has disease ($d$) | No disease ($\sim d$) |
| Acts | Treat ($t$) | 10 | 6 | 8 | 4 |
| | Not treat ($\sim t$) | 0 | 2 | 1 | 10 |

Section 2 describes a medical diagnosis (MD) problem, a small symmetric decision problem involving Bayesian revision of probabilities. Section 3 describes a strategy matrix representation and solution of the MD problem. Section 4 describes a decision tree representation and solution of the MD problem. Section 5 describes the scenario tree representation and the new pruning method. Section 6 describes a game tree representation of the MD problem and its solution using the rollback method and the pruning method. Section 7 describes a comparison of the different techniques. Finally, in section 8, we conclude.

## 2 A MEDICAL DIAGNOSIS PROBLEM

In this section, we give a statement of a medical diagnosis (MD) problem [Shenoy 1994]. The MD problem is a simple symmetric decision problem that involves Bayesian revision of probabilities. We will use this problem to illustrate the different representation and solution techniques.

A physician is trying to devise a policy for treating patients suspected of suffering from a disease $d$. $d$ causes a pathological state $p$ that in turn causes symptom $s$. The physician first observes whether or not a patient has symptom $s$. Based on this observation, she either treats the patient (for $d$ and $p$) or not. The physician's utility function depends on her decision to treat or not, the presence or absence of disease $d$, and the presence or absence of pathological state $p$. The prior probability of disease $d$ is 10%. For patients known to suffer from $d$, 80% suffer from pathological state $p$. On the other hand, for patients known not to suffer from $d$, 15% suffer from $p$. For patients known to suffer from $p$, 70% exhibit symptom $s$. And for patients known not to suffer from $p$, 20% have symptom $s$. Let variable D denotes the presence or absence of disease $d$, let variable P denotes the presence or absence of pathological state $p$, and let variable S denote the presence or absence of symptom $s$. We assume D and S are conditionally independent given P. Table I shows the physician's utility function.

## 3 STRATEGY MATRICES

In this section, we describe the strategy matrix representation and solution technique for decision problems. A strategy matrix representation of a decision problem is derived from von Neumann and Morgenstern's [1944] normal form representation of an $n$-person game.

The main task in the MD problem is to determine an optimal strategy. Since the physician observes the symptom before making a diagnosis, she has four distinct strategies as follows: $\sigma_1 = (t, t)$ meaning choose T = $t$ if S = $s$, and choose T = $t$ if S = $\sim s$ (i.e., choose $t$ regardless of the observed value of S), $\sigma_2 = (t, \sim t)$, $\sigma_3 = (\sim t, t)$, and $\sigma_4 = (\sim t, \sim t)$. There are eight possible events in the joint space of D, P, and S: $(d, p, s)$, $(d, p, \sim s)$, $(d, \sim p, s)$, $(d, \sim p, \sim s)$, $(\sim d, p, s)$, $(\sim d, p, \sim s)$, $(\sim d, \sim p, s)$, and $(\sim d, \sim p, \sim s)$. In a strategy matrix representation, each strategy constitutes a row of the matrix, each event constitutes a column, and for each strategy-event pair, there is a unique utility value determined by the utility function. For example, consider the pair $((t, \sim t), (\sim s, p, d))$. Since S = $\sim s$, the strategy $(t, \sim t)$ specifies T = $\sim t$. Accordingly, the utility associated with this pair is the utility value $\upsilon(\sim t, p, d) = 0$. Table II shows a strategy matrix representation of the MD problem. (In Table II, the expected utilities shown in bold are computed during the solution phase and are not part of the strategy matrix representation.)

The probabilities of all event are obtained by computing the joint probability distribution of all random variables in the problem (see left-hand side of Figure 1).

Solving a strategy matrix representation is straightforward. For each strategy, we compute its expected utility. Next, we identify an optimal strategy by identifying the maximum expected utility. In the MD problem, the expected utility of each strategy is shown in the last column in Table II. The maximum expected utility is 7.998. Thus the optimal strategy is (*t*, ~*t*), i.e., treat if S = *s*, and not treat if S = ~*s*.

**Table II**. A Strategy Matrix Representation and Solution of the MD Problem

| *UTILITIES* | | *EVENTS* $(s,p,d)$ | $(s,p,\sim d)$ | $(s,\sim p,d)$ | $(s,\sim p,\sim d)$ | $(\sim s,p,d)$ | $(\sim s,p,\sim d)$ | $(\sim s,\sim p,d)$ | $(\sim s,\sim p,\sim d)$ | *EU* |
|---|---|---|---|---|---|---|---|---|---|---|
| | $(t, t)$ | 10 | 6 | 8 | 4 | 10 | 6 | 8 | 4 | **4.8300** |
| *STRA-* | $(t, \sim t)$ | 10 | 6 | 8 | 4 | 0 | 2 | 1 | 10 | **7.9880** |
| *TEGIES* | $(\sim t, t)$ | 0 | 2 | 1 | 10 | 10 | 6 | 8 | 4 | **4.7820** |
| | $(\sim t, \sim t)$ | 0 | 2 | 1 | 10 | 0 | 2 | 1 | 10 | **7.9400** |
| *PROBABILITY* | | 0.0560 | 0.0945 | 0.0040 | 0.1530 | 0.0240 | 0.0405 | 0.0160 | 0.6120 | |

## 4 DECISION TREES

In this section, we describe a decision tree representation and solution of the MD problem. See Raiffa [1968] for a primer on the decision tree representation and solution method.

**Decision Tree Representation.** Figure 1 shows the preprocessing of probabilities that has to be done before we can complete a decision tree representation of the MD problem. In the probability tree on the left, we compute the joint probability distribution by multiplying the conditionals. For example, Pr(*d*, *p*, *s*) = Pr(*d*) Pr(*p* | *d*) Pr(*s* | *p*) = (.10)(.80)(.70) = .0560. In the probability tree on the right, we compute the desired conditionals by additions and divisions.

Figure 2 shows a complete decision tree representation of the *Medical Diagnosis* problem. (In Figure 2, the utility values shown in bold are computed during the solution phase and are not part of the representation.) Each path from the root node to a leaf node represents a *scenario*. This tree has 16 scenarios. The tree is *symmetric*, i.e., each scenario includes the same four variables S, T, P, and D, in the same sequence STPD.

**Rollback Method.** Starting from the leaves, we recursively delete all random and decision variable nodes in the tree. We delete each random variable node by averaging the utilities at the end of its edges with the probability distribution at that node ("averaging out"). We delete each decision variable node by maximizing the utilities at the end of its edges ("folding back"). This method is called *rollback*. The results of the rollback method for the MD problem are shown in Figure 2.

**Figure 2.** A Decision Tree Representation and Solution of the MD Problem

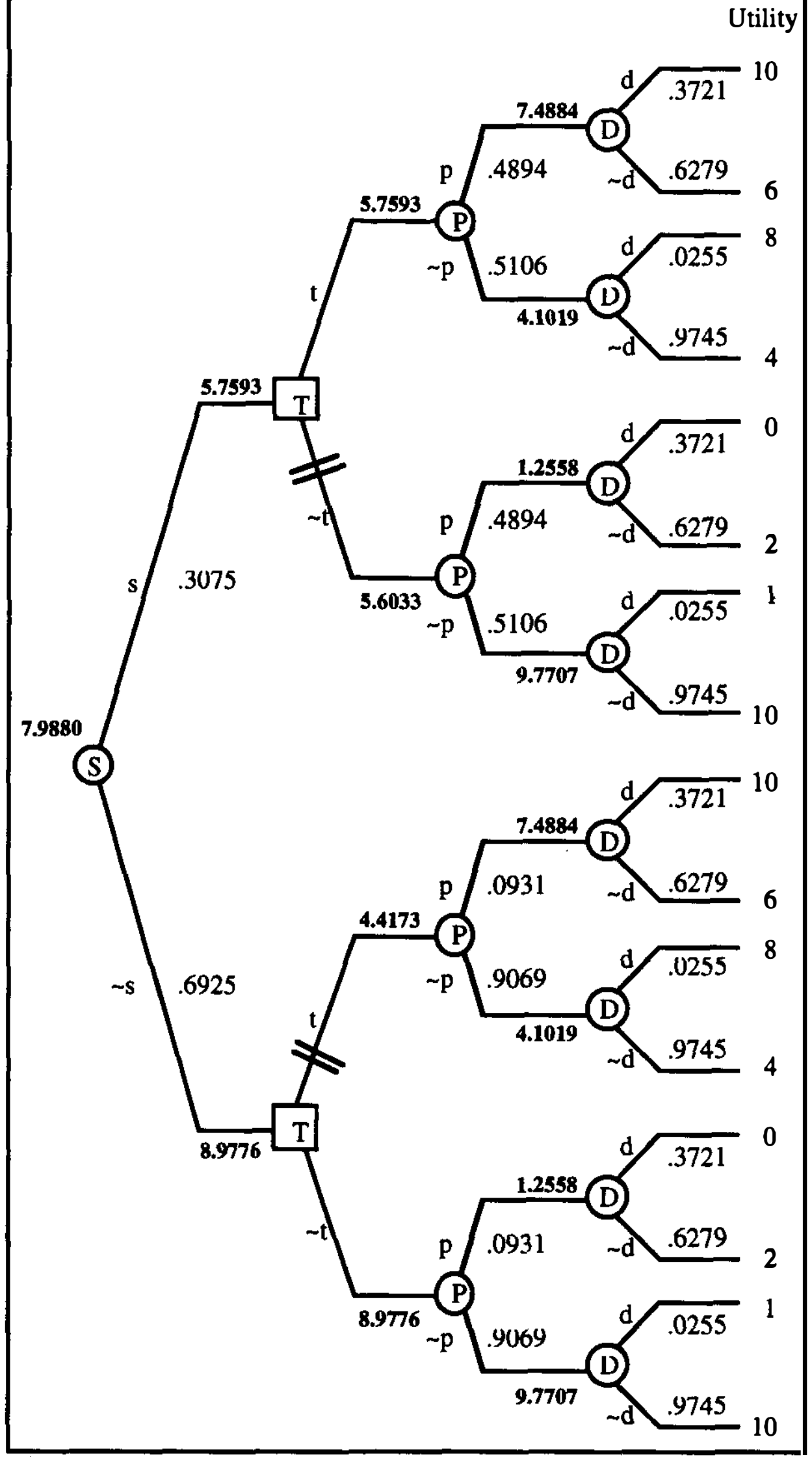

**Figure 1.** The Preprocessing of Probabilities in the MD Problem

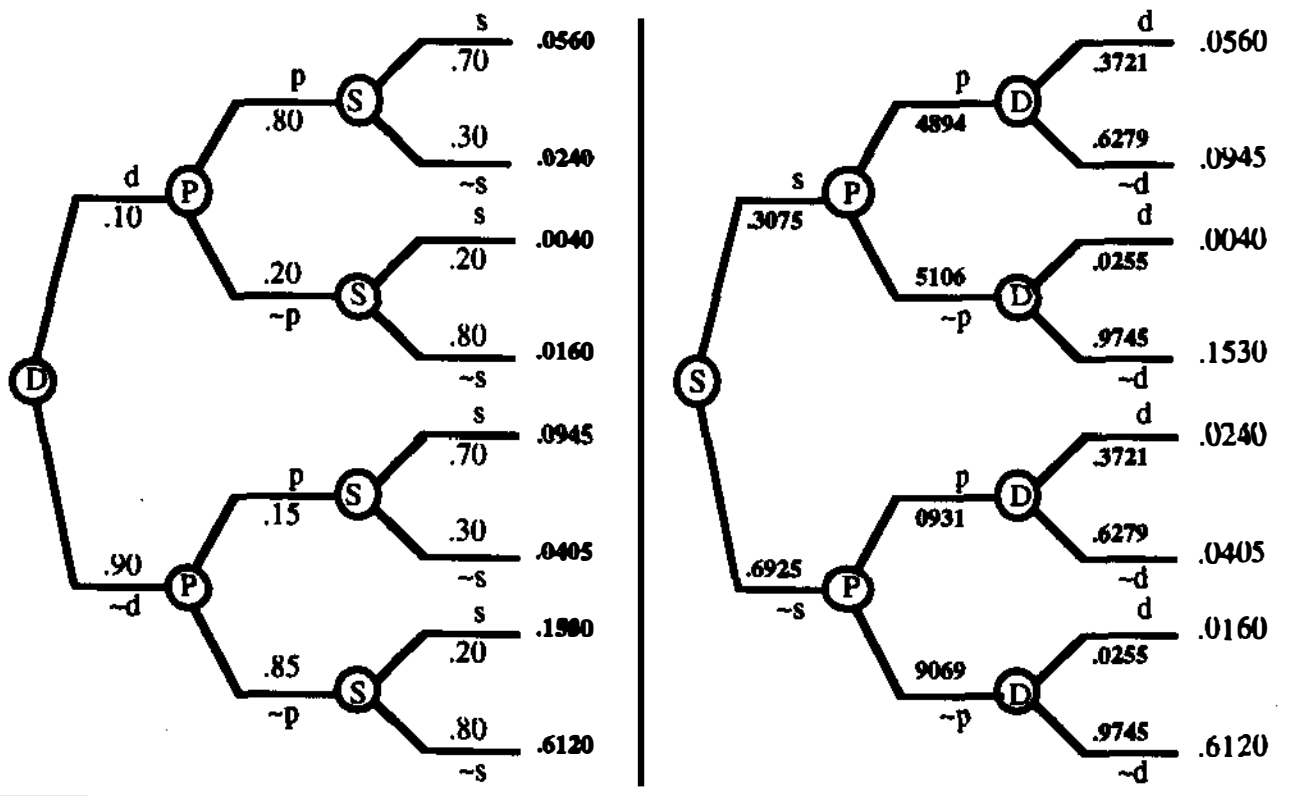

## 5 SCENARIO TREES

In this section, we describe the scenario tree representation and solution technique.

A scenario tree representation of the MD problem is shown in Figure 3. (In Figure 3, the utility values shown in bold are computed during the solution phase and are not part of the representation.) A scenario tree is slightly different from a decision tree. In a scenario tree, we do not have to specify probabilities for each edge emanating from a chance node. Instead we have to specify probabilities for each scenario, i.e., for each path from the root node to a leaf node. We call these probabilities *path probabilities*. For a leaf node, the path probability represents the conditional probability of reaching the leaf node conditional on a DM's strategy that makes the leaf node reachable. This probability is simply the conditional probability of all events on the path conditional on all acts on the path. For example, the probability of the top-most path in the scenario tree of Figure 3 is Pr(S = *s*, P = *p*, D = *d* | T = *t*).

**Figure 3.** A Scenario Tree Representation and Solution of the MD Problem

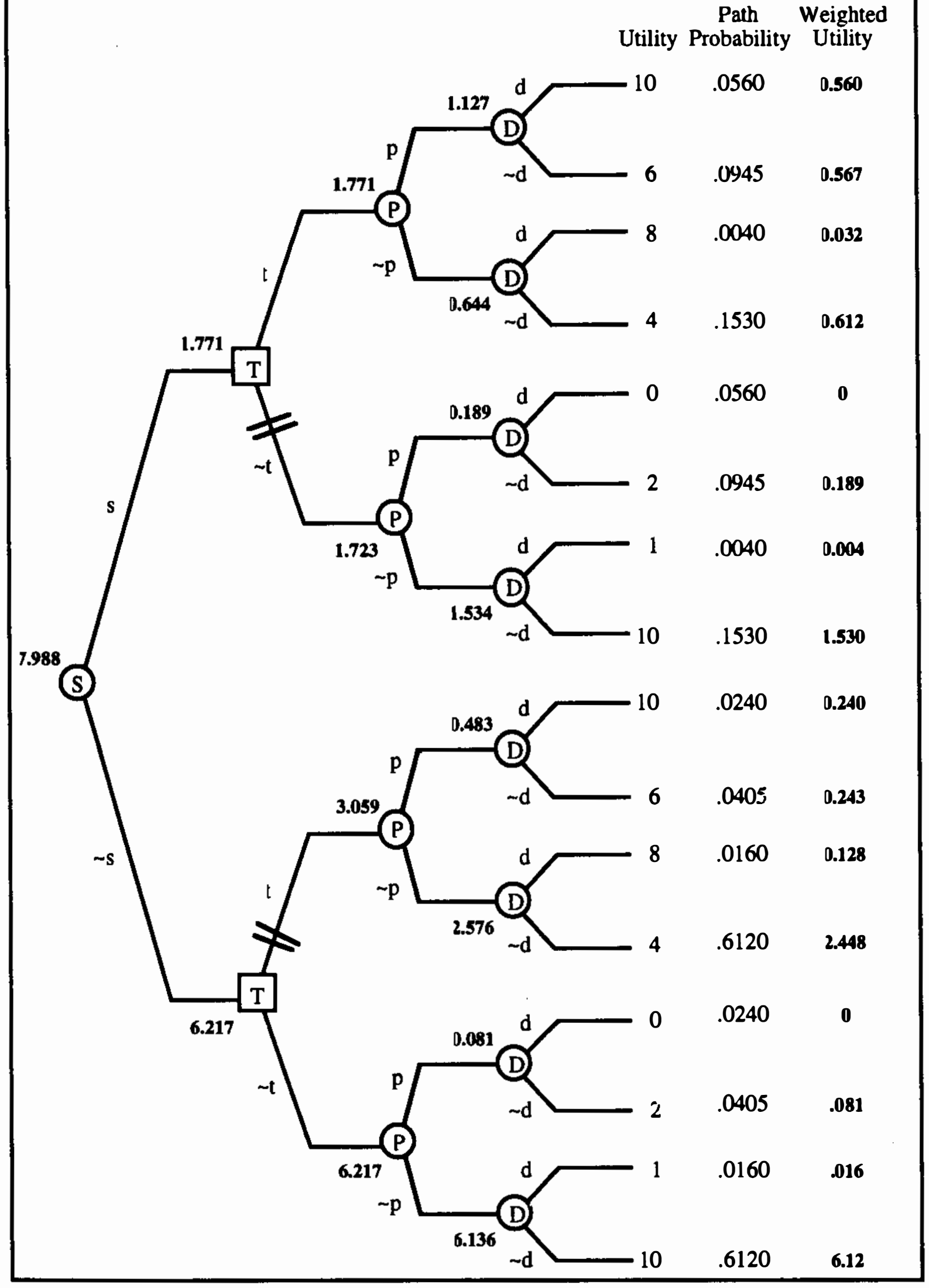

Let $\mathcal{S}$ denote the set of all scenarios. Consider the probability function $\pi$: $\mathcal{S} \to [0, 1]$ that assigns the path probability $\pi(\mathbf{x})$ to each scenario $\mathbf{x} \in \mathcal{S}$. We call the function $\pi$ the *path probability function.* Consider a strategy **y** available to the DM. We can encode **y** as a function $\xi_{\mathbf{y}}$: $\mathcal{S} \to \{0, 1\}$such that $\xi_{\mathbf{y}}(\mathbf{x}) = 0$ if scenario **x** has a zero probability of occurring, and $\xi_{\mathbf{y}}(\mathbf{x}) = 1$ otherwise. We call $\xi_{\mathbf{y}}$ a strategy function corresponding to strategy **y**. Given a strategy function $\xi_{\mathbf{y}}$, we can define the product $\pi \otimes \xi_{\mathbf{y}}$ as the function $\pi \otimes \xi_{\mathbf{y}}$: $\mathcal{S} \to [0, 1]$ such that $(\pi \otimes \xi_{\mathbf{y}})(\mathbf{x}) = \pi(\mathbf{x})\, \xi_{\mathbf{y}}(\mathbf{x})$ for each $\mathbf{x} \in \mathcal{S}$. A path probability function $\pi$ has the following property. If **y** is a strategy, then $\pi \otimes \xi_{\mathbf{y}}$ is a probability distribution function, i.e., $\Sigma\{(\pi \otimes \xi_{\mathbf{y}})(\mathbf{x}) \mid \mathbf{x} \in \mathcal{S}\} = 1$.

Let $\upsilon$: $\mathcal{S} \to \mathbb{R}$ denote the utility function, i.e., $\upsilon(\mathbf{x})$ is the DM's utility under scenario **x**. Consider the product $\pi \otimes \upsilon$: $\mathcal{S} \to \mathbb{R}$ defined as follows: $(\pi \otimes \upsilon)(\mathbf{x}) = \pi(\mathbf{x})\,\upsilon(\mathbf{x})$ for all $\mathbf{x} \in \mathcal{S}$. We call $\pi \otimes \upsilon$ the *weighted utility function.*

If the DM chooses strategy **y**, then the DM's expected utility is $\Sigma\{((\pi \otimes \xi_{\mathbf{y}}) \otimes \upsilon)(\mathbf{x}) \mid \mathbf{x} \in \mathcal{S}\}$. Notice that $(\pi \otimes \xi_{\mathbf{y}}) \otimes \upsilon = (\pi \otimes \upsilon) \otimes \xi_{\mathbf{y}}$. Thus the problem can be stated as follows: Find strategy **y** so as to maximize $\Sigma\{((\pi \otimes \upsilon) \otimes \xi_{\mathbf{y}})(\mathbf{x}) \mid \mathbf{x} \in \mathcal{S}\}$. The pruning method that we will now describe solves precisely this problem using deterministic dynamic programming.

**The Pruning Method.** First, for each leaf node, we compute the weighted utility by multiplying the utility and the path probability. Hence forth, we use the weighted utilities for pruning chance and decision nodes.

**Pruning Chance Nodes.** If we have a chance node all of whose edges end in leaf nodes, first we compute the sum of the weighted utilities at the end of its edges, and next we replace the subtree associated with the chance node by a payoff node whose value is set equal to the computed sum of weighted utilities.

**Pruning Decision Nodes.** If we have a decision node all of whose edges end in payoff nodes, first we compute the maximum of the weighted utilities, and next we replace the subtree associated with the decision node by a payoff node whose value is set at the computed maximum value of weighted utilities. Each time we prune a decision node, we keep track of a value of the decision node where the maximum is achieved.

When we have pruned all decision nodes, we have an optimal strategy. When we have pruned all decision and chance nodes, we end with one payoff node whose utility value is the expected utility of an optimal strategy. This pruning method is illustrated in Figure 3.

## 6 GAME TREES

The game tree representation technique for decision problems is described in [Shenoy 1993b]. It is based on von Neumann and Morgenstern's [1944] extensive form representation of *n*-person games. Figure 4 shows a game tree representation of the MD problem. (In Figure 4, the utilities shown in bold are computed during the solution phase and are not part of the game tree representation.)

Game trees are similar in many ways to decision trees. The main difference between the two representation techniques is the encoding of information constraints. In decision tree, information constraints are encoded in the sequencing of chance and decision nodes in each scenario. Thus if the value of a chance variable R is known to the decision maker when she chooses a value of a decision variable N, then R must precede N on the paths from the root node to leaf nodes. In game trees, information constraints are encoded by information sets. Information sets are a partition of the set of decision nodes in a game tree. When faced with a decision, a decision maker knows which information set she is in, but does not know which node in an information set she is at. In Figure 4, there are two information sets $I_1$ and $I_2$ each containing 4 instances of node T. In information set $I_1$, the physician knows that S = *s*, but not the values of D and P. In information set $I_2$, the physician knows that S = ~*s*, but not the values of D and P. In game trees, the sequence of variables on a path from the root node to a leaf node need not represent information constraints. Instead, the sequence may represent time, causation, etc. If we have a causal probability model (as we do in the MD problem), and we use the sequence to represent causation, then no preprocessing is required to represent a decision problem as a game tree. In Figure 4, for example, all the probabilities on edges are specified in the statement of the MD problem.

**Figure 4**. A Game Tree Representation and Solution of the MD Problem

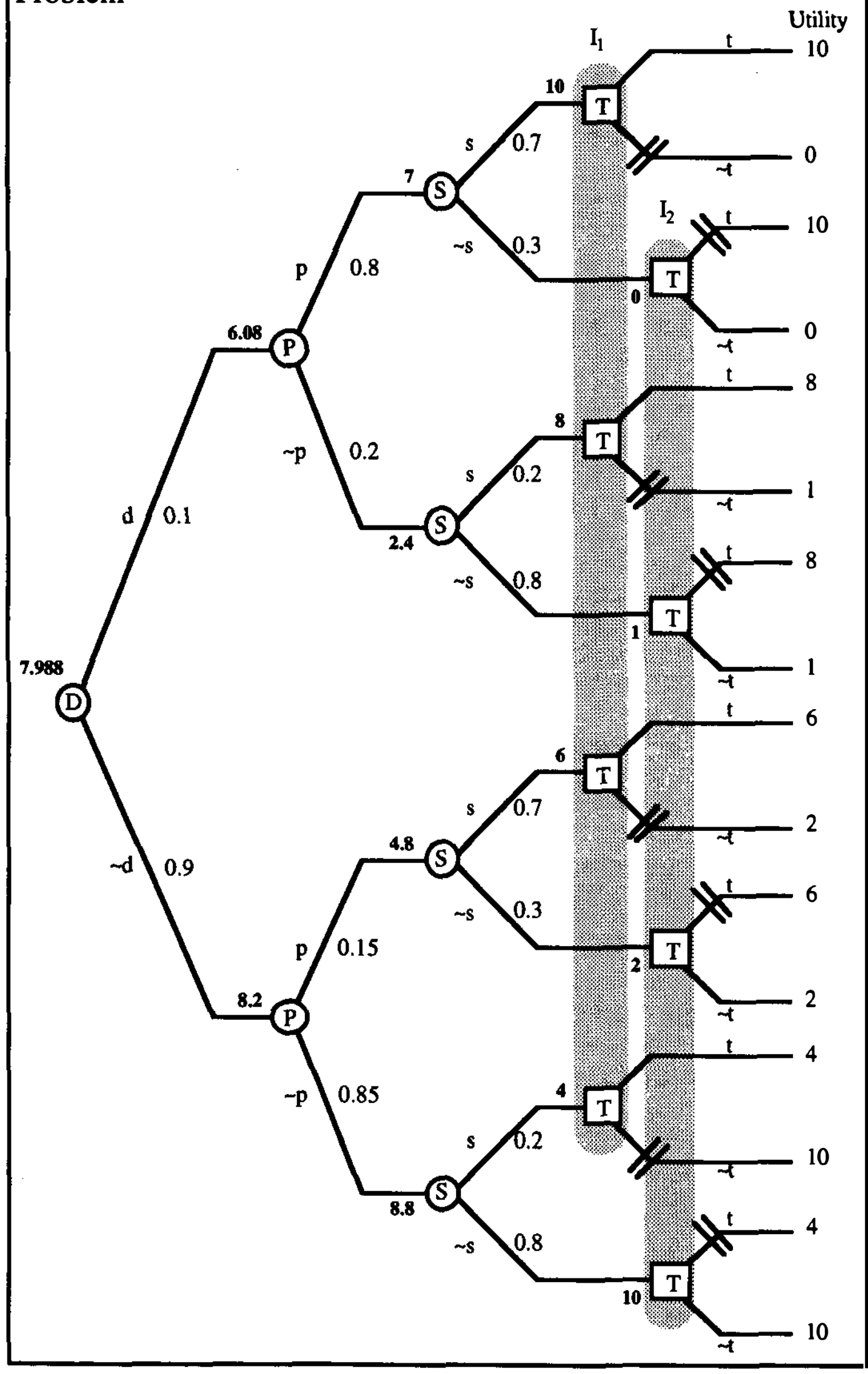

**The Rollback Method for Game Trees**. The decision tree rollback method generalizes to game trees [Shenoy 1993b]. We start from the neighbors of payoff nodes and go toward the root until the entire tree has been pruned. The rule for pruning chance nodes is exactly the same as in the rollback procedure of decision trees. The rule for pruning decision nodes is slightly different from the rollback procedure of decision trees.

Suppose we have an information set such that the edges leading out of the decision nodes end at payoff nodes. First, we compute the conditional probability distribution on the decision nodes of the information set conditioned on the event that we have reached the information set (the details of this step are explained in the following paragraph). Second, for each value of the decision variable associated with the information set, we compute the expected payoff using the payoffs at the end of corresponding edges and using the conditional distribution computed in the first step. Third, we identify a value of the decision variable (and the corresponding edges of each decision node) associated with the maximum expected payoff. Fourth, we prune each decision node by replacing the corresponding subtree by a payoff node whose payoff is equal to the payoff at the end of its edge identified in step 3. We call this technique (for pruning decision nodes in an information set) pruning by *maximization of conditional expectation*. Pruning by maximization of conditional expectation generalizes the method of pruning a singleton information set by maximization.

Computing the conditional distribution for an information set is easy. For each decision node in the information set, we simply multiply all probabilities on the chance edges in the path from the root to the decision node. This gives an unconditional distribution on the nodes in the information set. The sum of these probabilities gives us the probability of reaching the information set assuming prior decisions that allow us to get there. To compute the conditional distribution, we normalize this unconditional distribution by dividing by the sum of the probabilities. From a computational perspective, the normalization step is unnecessary and can be dispensed with.

We will illustrate pruning decision nodes by conditional expectation for the game tree representation of the MD problem. Consider information set $I_1$ in Figure 4. Notice that all four decision nodes in $I_1$ have edges leading to payoff nodes. The unnormalized distribution for the three nodes in $I_1$ (from top to bottom) is given by the probability vector (0.1*0.8*0.7, 0.1*0.2*0.2, 0.9*0.15*0.7, 0.9*0.85*0.2) = (0.0560, 0.0040, 0.0945, 0.1530). Thus for T = *t*, the (unnormalized) expected payoff is (0.0560 * 10) + (0.0040*8) + (0.0945*6) + (0.1530*4) = 1.771, and for T = ~*t*, the expected payoff is (0.0560*0) + (0.0040*1) + (0.0945*2) + (0.1530*10) = 1.723. Since 1.771 > 1.723, we identify edge T = *t* with each node in $I_1$. The rollback method is illustrated in Figure 4.

**The Pruning Method for Game Trees.** Next, we describe a generalization of the pruning method described in the previous section so that it applies to game trees. First, we compute path probabilities and weighted utilities for all payoff nodes. The rule for pruning chance nodes is exactly the same as in decision trees. The rule for pruning decision nodes is as follows.

**Pruning Decision Nodes.** Suppose we have an information set such that the edges leading out of the decision nodes end at payoff nodes. First, for each value of the decision variable, we compute the sum of the weighted utilities at the end of the respective edges of the decision nodes in the information set. Second, we identify a value of the decision variable that has the maximum sum of weighted utilities computed in the first step. Third, we prune each decision node by replacing the corresponding subtree by a payoff node whose value is equal to the weighted utility at the end of its edge corresponding to the value of the decision variable identified in the second step.

We will illustrate the method for pruning decision nodes in game trees using the MD problem. Consider decision nodes in information set $I_1$. For T = *t*, the sum of the weighted utilities is 0.560 + 0.032 + 0.567 + 0.612 = 1.771, and for T = ~*t*, the sum of the weighted utilities is 0 + .004 + 0.189 + 1.530 = 1.723. Since 1.771 > 1.723, we identify edge T = *t* with each node in $I_1$. The results of executing the pruning method for the MD game tree are shown in Figure 5.

The correctness of this method follows from exactly the same arguments as in the case of scenario trees. The computational efficiencies of the rollback method and the pruning method are compared in the next section.

Figure 5. The MD Game Tree Solved Using the Pruning Method

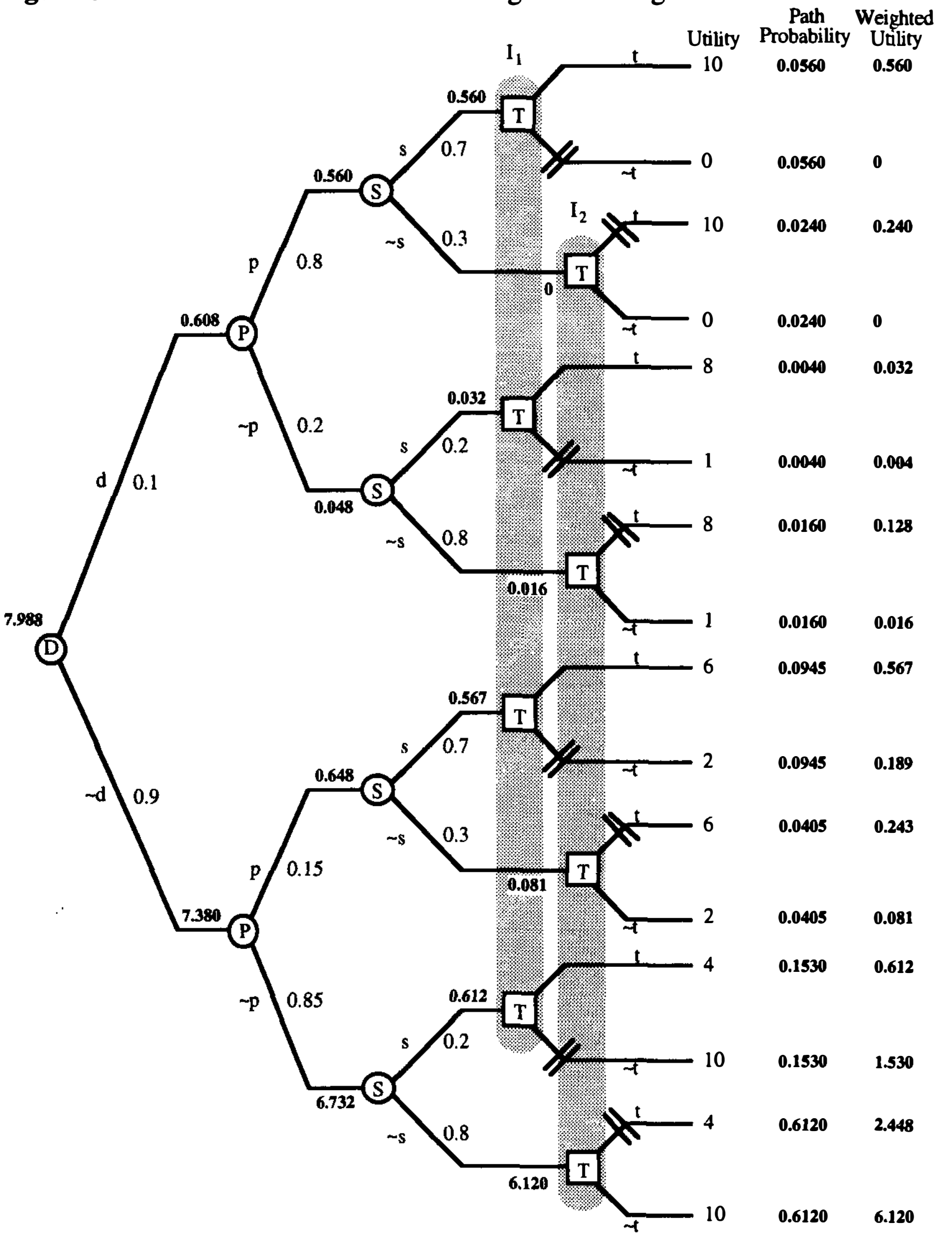

## 7 COMPARISON

In this section, we compare scenario trees and game trees to strategy matrices and decision trees. Also we compare the pruning method of scenario trees and game trees to the rollback method of decision trees and game trees.

**Strategy Matrices.** The strategy matrix representation involves considerable preprocessing. Not only do we have to compute the joint probability distribution of all chance variables, we also have to identify all possible strategies. The number of configurations in the frame of all chance variables is an exponential function of the number of chance variables, and the number of strategies is an exponential function of the number of decision nodes in a decision tree. Thus, representation and solution of strategy matrices involve global computation (on the space of configurations of all chance variables) to compute the joint probability distribution, and global computation (on the space of all strategies) to compute an optimal strategy. Strategy matrices have some advantages. They can be used to define optimal strategies. Also, the solution method of strategy matrices yields not only the utility of an optimal strategy but also the utilities of all strategies. For example, in the MD problem, the strategy (~*t*, ~*t*) has a utility of 7.940 that is only marginally lower than the utility of the optimal strategy.

**Figure 6.** Computing the Conditional Probability Distributions Using the Arc Reversal Method

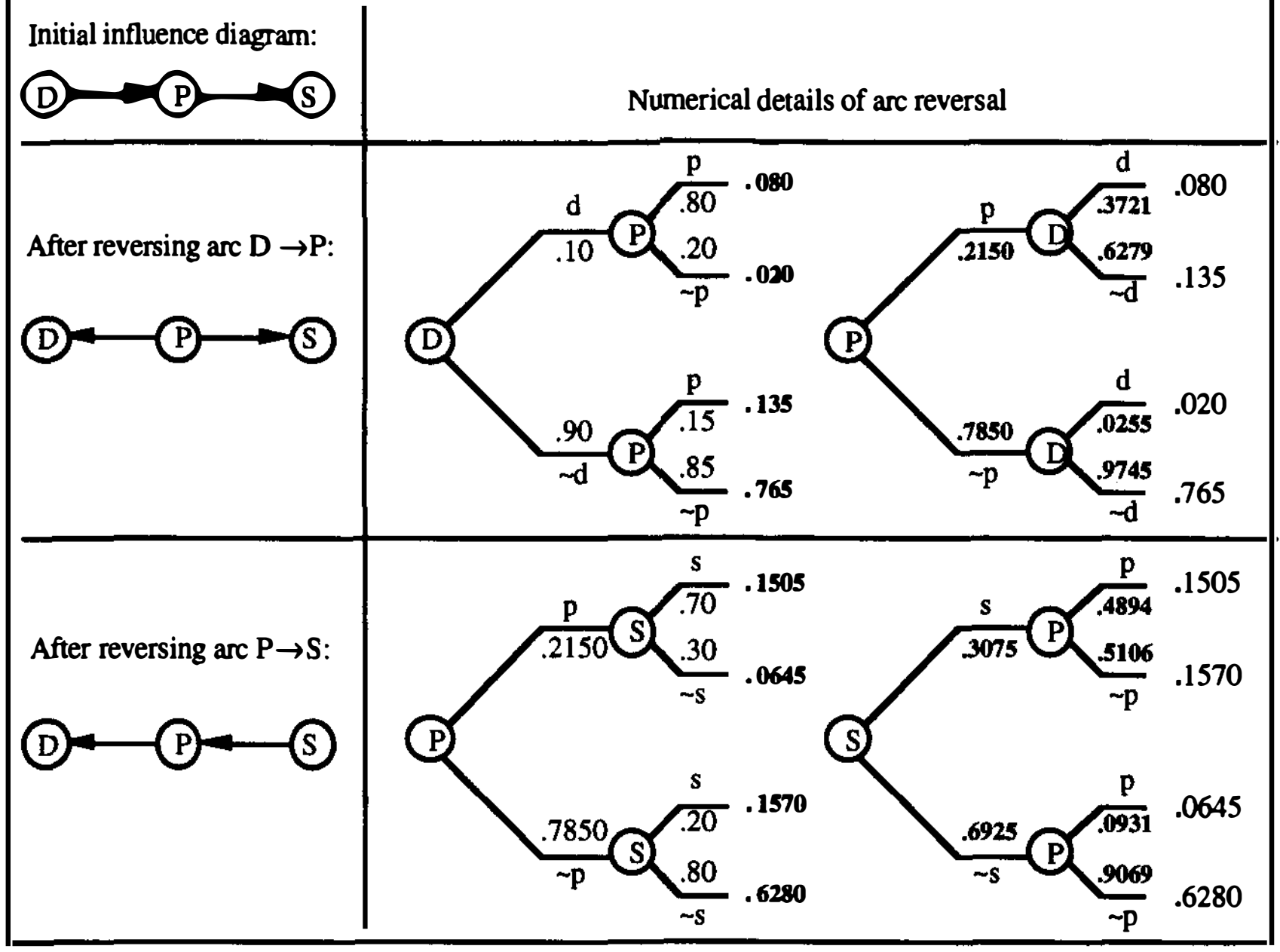

In the MD problem, representation and solution of the strategy matrix (shown in Table II) involves a total of 75 operations (where we count each addition, multiplication, division, and comparison as an operation). A breakdown is as follows. Computing the joint probability distribution involves 12 operations, and computing the maximum expected utility involves 63 operations (15 operations to compute the expected utility of each strategy and 3 operations to identify the maximum).

In general, suppose we have a symmetric decision problem with *m* chance variables each of which has 2 values, with *n* decision variables each of which has 2 values, and with $2^k$ distinct strategies[1]. Then representing and solving the problem using the strategy matrix method requires a total of approximately $2^{m+k+1} + 2^{m+1}$ operations (approximately $2^{m+1}$ to compute the joint probability distribution, $2^{m+1} - 1$ to compute the expected utility of each strategy, and approximately $2^k$ to identify an optimal strategy).

**Decision Trees.** The decision tree representation requires conditional probability distributions for each chance node in a decision tree. If we compute the conditional probability distributions from the joint probability distribution (as described in Figure 1), then this involves global computation on the space of all configurations. The rollback method however computes an optimal strategy using local computation.

In the MD problem, representation and solution of the decision tree (shown in Figure 2) involve a total of 71 operations. A breakdown of the total is as follows. Computing the conditional probability distributions as shown in Figure 1 involves 30 operations, and executing the rollback method as shown in Figure 2 involves 41 operations (3 operations for each of the chance nodes and 1 operation for each of the decision nodes).

The decision tree representation and solution technique can be made more efficient as follows. First, we can use the arc-reversal method [Olmsted 1983, Shachter 1986] to compute the conditional probability distributions. The arc-reversal method uses local computation. This is illustrated in Figure 6. Only 20 operations are needed to compute the requisite conditionals using the arc-reversal method. Second, the decision tree solution can be made more efficient by identifying coalescence [Olmsted 1983]. In the MD problem, by using coalescence, we can simplify the decision tree as shown in Figure 7. The rollback method can be executed for the coalesced decision tree using only 23 operations (3 operation for each chance node and 1 operation for each decision node). Thus by using the arc-reversal method and coalescence, we can represent and solve the MD problem using only 43 operations. We are assuming here that identifying coalescence is done during the representation phase by the modeler.

In general, automating coalescence in decision trees is not easy. However, coalescence is automatically achieved in influence diagrams and valuation networks [Shenoy 1994]. In influence diagrams and valuation networks, coalescence is achieved by local computation. The arc-reversal method of influence diagrams solves the MD problem using 43 operations, and the fusion algorithm of valuation networks [Shenoy 1992, 1993a] solves the MD problem using 31 operations [Shenoy 1994].

**Scenario Trees.** Scenario trees do not require conditional probability distributions. However, we do require path probabilities. In the MD problem, computation of the path probabilities is achieved by computing the joint distribution of all chance variables. This involves 12 operations. Executing the pruning method for the scenario tree representation shown in Figure 3 involves a total of 31 operations (16 operations to compute the weighted utilities, and 1 operation per decision and chance node). Thus a total of 43 operations are required to represent and solve the MD problem using the scenario tree method.

[1] *k* denotes, for example, the number of decision nodes in a decision tree representation, or the number of information sets in a game tree representation. The value of *k* depends on the information constraints. It can be easily shown that $2^n - 1 \leq k \leq 2^{m+n} - 2^m$.

**Figure 7.** A Decision Tree Representation and Solution of the MD Problem Using Coalescence

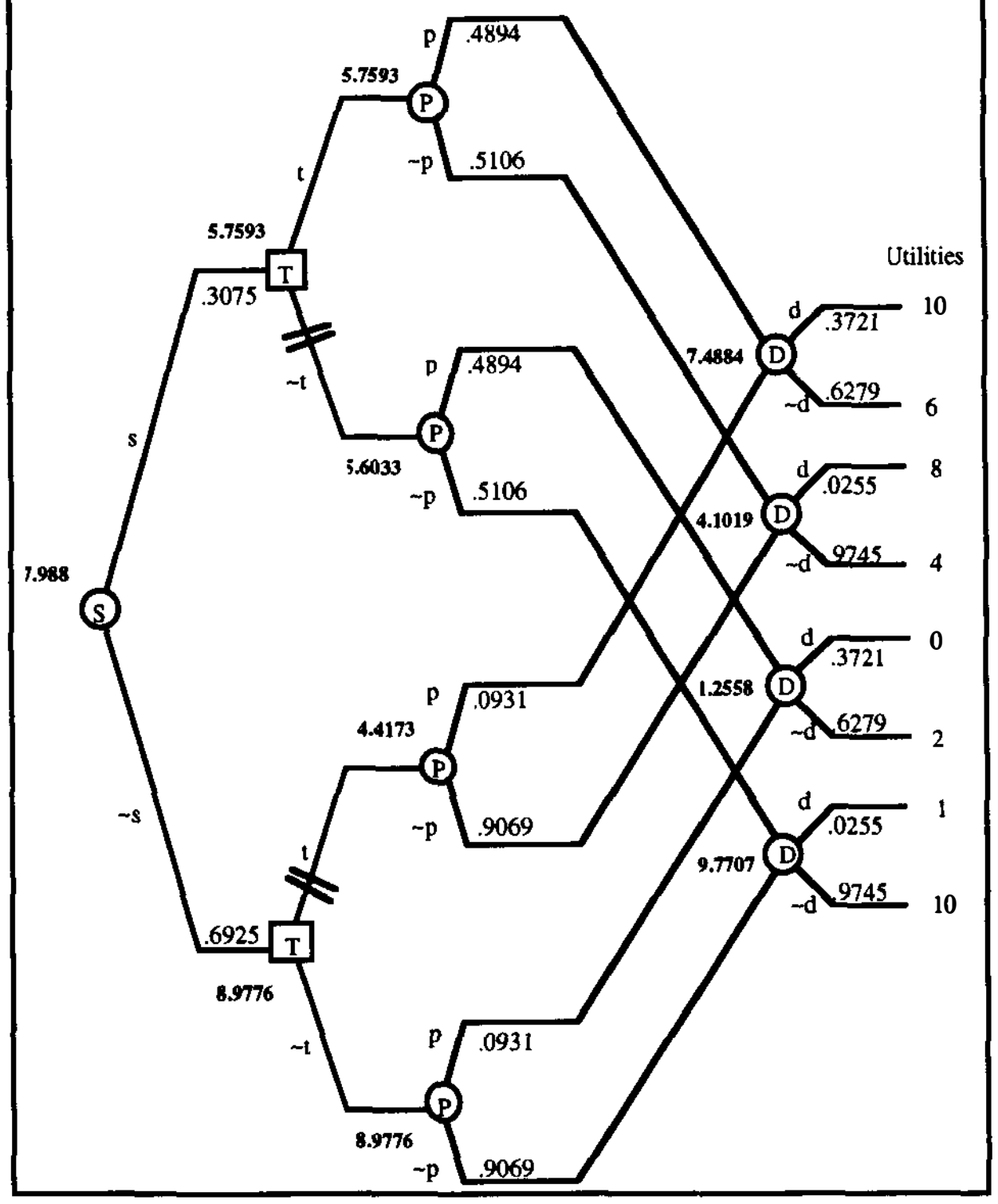

In general, the scenario tree method involves working on the global space of all configurations. Unlike decision trees, we cannot use either arc-reversal or coalescence to make the representation and solution more efficient. The pruning method of scenario trees does use local computation to identify an optimal strategy and compute its utility.

Suppose that we have a symmetric decision problem with $m$ chance variables and $n$ decision variables, suppose that each variable has two values, and suppose that all conditional probabilities required in the decision tree representation are specified in the problem. A decision tree representation would have a total of $2^{m+n} - 1$ decision and chance nodes, and $2^{m+n}$ leaf nodes. Suppose that the decision tree representation has approximately the same number of decision and chance nodes, say $2^{m+n-1}$ chance nodes and $2^{m+n-1}$ decision nodes. Then the rollback method would require a total of approximately $2^{m+n+1}$ operations (3 operations to prune each chance node and 1 operation to prune each decision node), and the pruning method would require a total of approximately $2^{m+n+1}$ + $2^{m+1}$ operations ($2^{m+1}$ to compute the joint probability distribution, $2^{m+n}$ to compute the weighted utilities, 1 operation to prune each chance node and 1 operation to prune each decision node). In this case, the rollback method is more efficient than the pruning method.

Now suppose that we have the same decision problem as above except that Bayesian revision of probabilities is required before the problem can be represented as a decision tree. In this case, if we compute the required conditionals by computing the joint, this would require a total of approximately $2^{m+n+1} + 2^{m+2} + 2^{m}$ operations ($2^{m+1}$ operations to compute the joint, $2^{m+1} + 2^{m}$ operations to compute the required conditionals, and $2^{m+n+1}$ operations for rollback). If we use a scenario tree representation with the pruning method, then we would need a total of approximately $2^{m+n+1} + 2^{m+1}$ operations ($2^{m+1}$ to compute the joint probability distribution, $2^{m+n}$ to compute the weighted utilities, and $2^{m+n}$ operations to prune the nodes). In this case, the scenario tree representation with the pruning method is more efficient than the decision tree with the rollback method.

**Game Trees.** If we assume that we have a causal probability model for all chance variables, then unlike strategy matrices, decision trees, and scenario trees, game trees do not require any preprocessing. Executing the rollback method for the MD game tree (see Figure 4) requires 63 operations (12 to compute conditional probabilities of reaching decision nodes in information sets, 15 operation to prune each information set, and 3 operations to prune each chance node). In comparison, executing the pruning method for the MD game tree (see Figure 5) requires only 49 operations (12 operations to compute the path probabilities, 16 operations to compute the weighted utilities, 7 operations to prune the decision nodes in each of 2 information sets, and 1 operation to prune each of 7 chance nodes).

In general, suppose we have a game tree with $m$ chance variables each of which has 2 values, with $n$ decision variables each of which has 2 values, with $2^{m+n-1}$ chance nodes, and with $2^{m+n-1}$ decision nodes in $k$ information sets. If we use the rollback method to solve the game tree, we would need a total of approximately $2^{m+n+1} + 2^{m+n} + 2^{m+n-1} + 2^{m+1}$ operations ($2^{m+1}$ operations to compute the unnormalized conditional probability distributions at each information set, $2^{m+n+1}$ to prune the $k$ information sets, and $3(2^{m+n-1})$ operations to prune the $2^{m+n-1}$ chance nodes). If we use the pruning method described in this paper, then we need a total of approximately $2^{m+n+1} + 2^{m+n-1} + 2^{m+1}$ operations ($2^{m+1}$ operations to compute the path probabilities, $2^{m+n}$ operations to compute the weighted utilities, $2^{m+n}$ to prune the $k$ information sets, and $2^{m+n-1}$ operations to prune the $2^{m+n-1}$ chance nodes). Thus the pruning method is more efficient than the rollback method for game trees.

**Table III**. A Comparison of Computational Efficiency of Different Techniques for the MD Problem

| *# Operations (+, ×, ÷, ≥) in MD Problem* | *Representation (Preprocessing)* | *Solution* | *TOTAL* |
|---|---|---|---|
| Strategy Matrix with Expected Values | 12 | 63 | 75 |
| Decision Tree with Rollback | 30 | 41 | 71 |
| Scenario Tree with Pruning | 12 | 31 | 43 |
| Game Tree with Rollback | 0 | 63 | 63 |
| Game Tree with Pruning | 0 | 49 | 49 |
| Influence Diagram with Arc Reversal | 0 | 49 | 49 |

**Table IV**. A Comparison of Computational Efficiencies of Different Techniques

| *Problem Characteristics* | *Representation and Solution Technique* | *Approximate Total # Operations (+, ×, ÷, ≥)* |
|---|---|---|
| General | Strategy Matrix with Expected Values | $2^{m+k+1} + 2^{m+1}$ |
| No Bayesian Revision | Decision Tree with Rollback | $2^{m+n+1}$ |
| No Bayesian Revision | Scenario Tree with Pruning | $2^{m+n+1} + 2^{m+1}$ |
| Bayesian Revision | Decision Tree with Rollback | $2^{m+n+1} + 2^{m+2} + 2^{m}$ |
| Bayesian Revision | Scenario Tree with Pruning | $2^{m+n+1} + 2^{m+1}$ |
| General | Game Tree with Rollback | $2^{m+n+1} + 2^{m+n} + 2^{m+n-1} + 2^{m+1}$ |
| General | Game Tree with Pruning | $2^{m+n+1} + 2^{m+n-1} + 2^{m+1}$ |

Table III summarizes the computational efficiencies of the different techniques for the MD problem, and Table IV summarizes the computational efficiencies of the different techniques in general

## 8 CONCLUSION

The main objective of this paper was to introduce a new pruning method to solve decision trees and game trees. In decision problems requiring Bayesian revision of probabilities for a decision tree representation, the new pruning method suggests a slight variant of decision trees that we call scenario trees. For such problems, the scenario tree representation with the pruning method is more efficient than the decision tree representation with the rollback method (assuming we compute the conditionals from the joint). The pruning method generalizes to game trees and it is more efficient than the rollback method for game trees.

## Acknowledgments

I am grateful to Rui Guo for comments and discussions.